\pdfoutput=1

\documentclass[11pt]{article}

\usepackage{EMNLP2022}
\usepackage{graphicx}
\usepackage{multirow}
\usepackage{times}
\usepackage{latexsym}
\usepackage{amsmath}
\usepackage{amssymb}
\usepackage{booktabs}

\usepackage[T1]{fontenc}

\usepackage[utf8]{inputenc}

\usepackage{microtype}

\usepackage{inconsolata}

\title{Align Vision-Language Semantics by Multi-task Learning for Multi-Modal Summarization}

\author{Chenhao Cui\textsuperscript{2}\footnotemark[1], Xinnian Liang\textsuperscript{1}\footnotemark[1], Shuangzhi Wu\textsuperscript{2}, Zhoujun Li\textsuperscript{1}\footnotemark[2]\\ 
\textsuperscript{1}State Key Lab of Software Development Environment, Beihang University, Beijing, China \\ 
\textsuperscript{2}Tencent Inc., Beijing, China\\ 
\texttt{\{xnliang,cuich,lzj\}@buaa.edu.cn,\{frostwu\}@tencent.com}
}

\begin{document}
\maketitle
\renewcommand{\thefootnote}{\fnsymbol{footnote}} 
\footnotetext[1]{Contribution Equally.} 
\footnotetext[2]{Corresponding Authors.} 
\renewcommand{\thefootnote}{\arabic{footnote}} 

\begin{abstract}
Most current multi-modal summarization methods follow a cascaded manner, where an off-the-shelf object detector is first used to extract visual features. After that, these visual features are fused with language representations for the decoder to generate the text summary. However, the cascaded way employs separate encoders for different modalities, which makes it hard to learn the joint vision and language representation. In addition, they also ignored the semantics alignment between paragraphs and images for multi-modal summarization tasks, which are crucial to a precise summary. To tackle these issues, in this paper, we propose ViL-Sum to jointly model paragraph-level \textbf{Vi}sion-\textbf{L}anguage Semantic Alignment and Multi-Modal \textbf{Sum}marization. Our ViL-Sum contains two components for better learning multi-modal semantics and aims to align them. The first one is a joint multi-modal encoder. The other one is two well-designed tasks for multi-task learning, including image reordering and image selection. Specifically, the joint multi-modal encoder converts images into visual embeddings and attaches them with text embedding as the input of the encoder. The reordering task guides the model to learn paragraph-level semantic alignment and the selection task guides the model to select summary-related images in the final summary. Experimental results show that our proposed ViL-Sum significantly outperforms current state-of-the-art methods. In further analysis, we find that two well-designed tasks and a joint multi-modal encoder can effectively guide the model to learn reasonable paragraph-image and summary-image relations.

\end{abstract}

\section{Introduction}
\begin{figure*}[ht]
\centering
\includegraphics[width=0.8\textwidth]{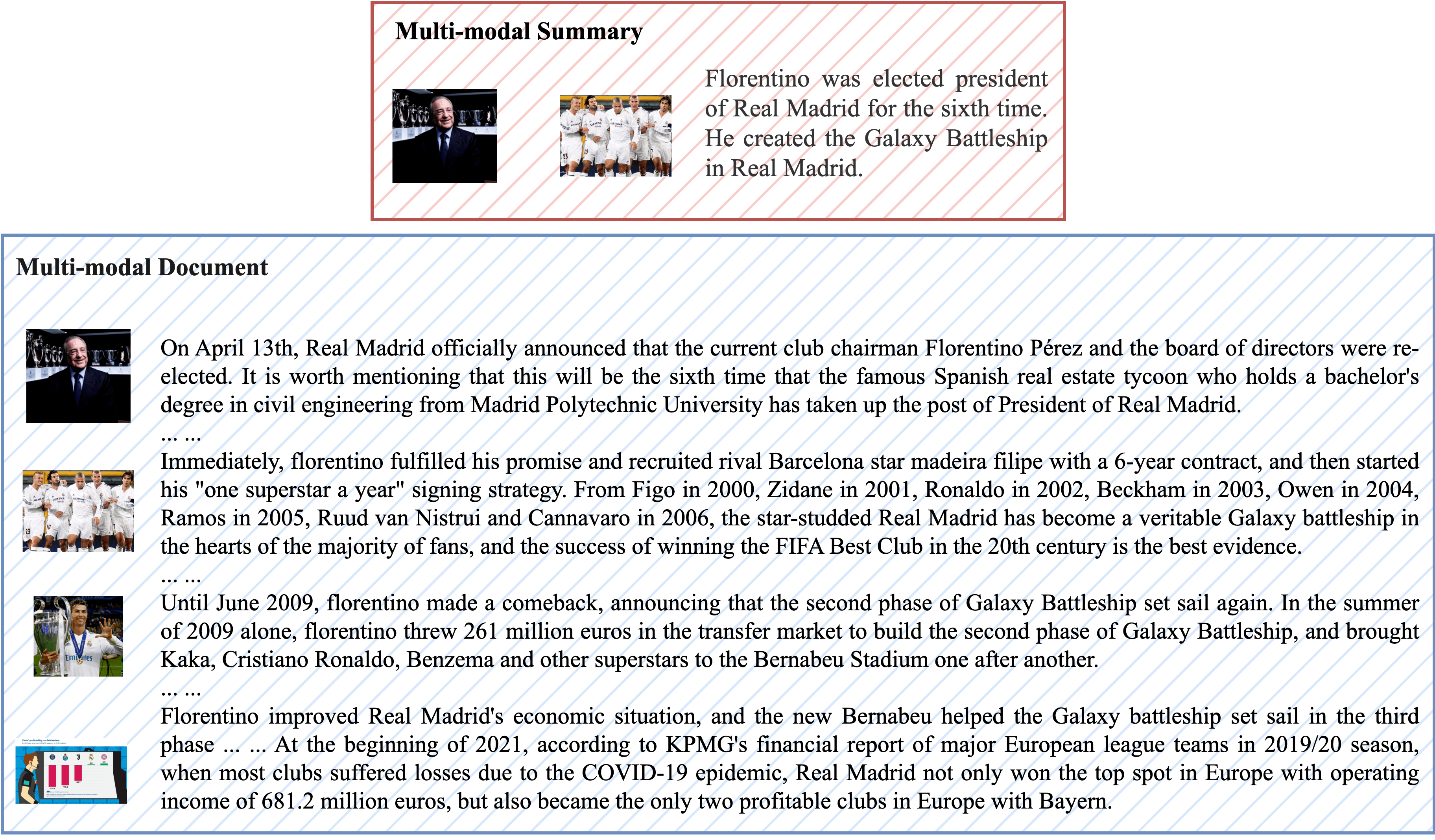}
\caption{An example for explaining the semantic alignment between images and paragraphs in the document. ``...'' means some content was omitted. Each image is aligned to the paragraph at right.}
\label{fig:example}
\end{figure*}

The dramatic increase of multi-modal data (including text, image, audio, and video) on the Internet makes research on multi-modal summarization necessary. Multi-modal summarization aims to generate a condensed summary, which can cover salient information from one or more modalities inputs~\cite{6527322,li-etal-2017-multi}. Different from traditional pure text summary, Zhu \textit{et.al.}~\cite{zhu-etal-2018-msmo} pointed out that generated summaries with both text and images can effectively increase the satisfaction of users. Intuitively, people can grasp key information easier from multiple modalities than only from the text. This task is defined as multi-modal summarization with multi-modal outputs (MSMO). Figure~\ref{fig:example} shows an example of this task, which gets text and images as input and generates one summary with two selected images.

\begin{figure}
    \centering
    \includegraphics[width=0.45\textwidth]{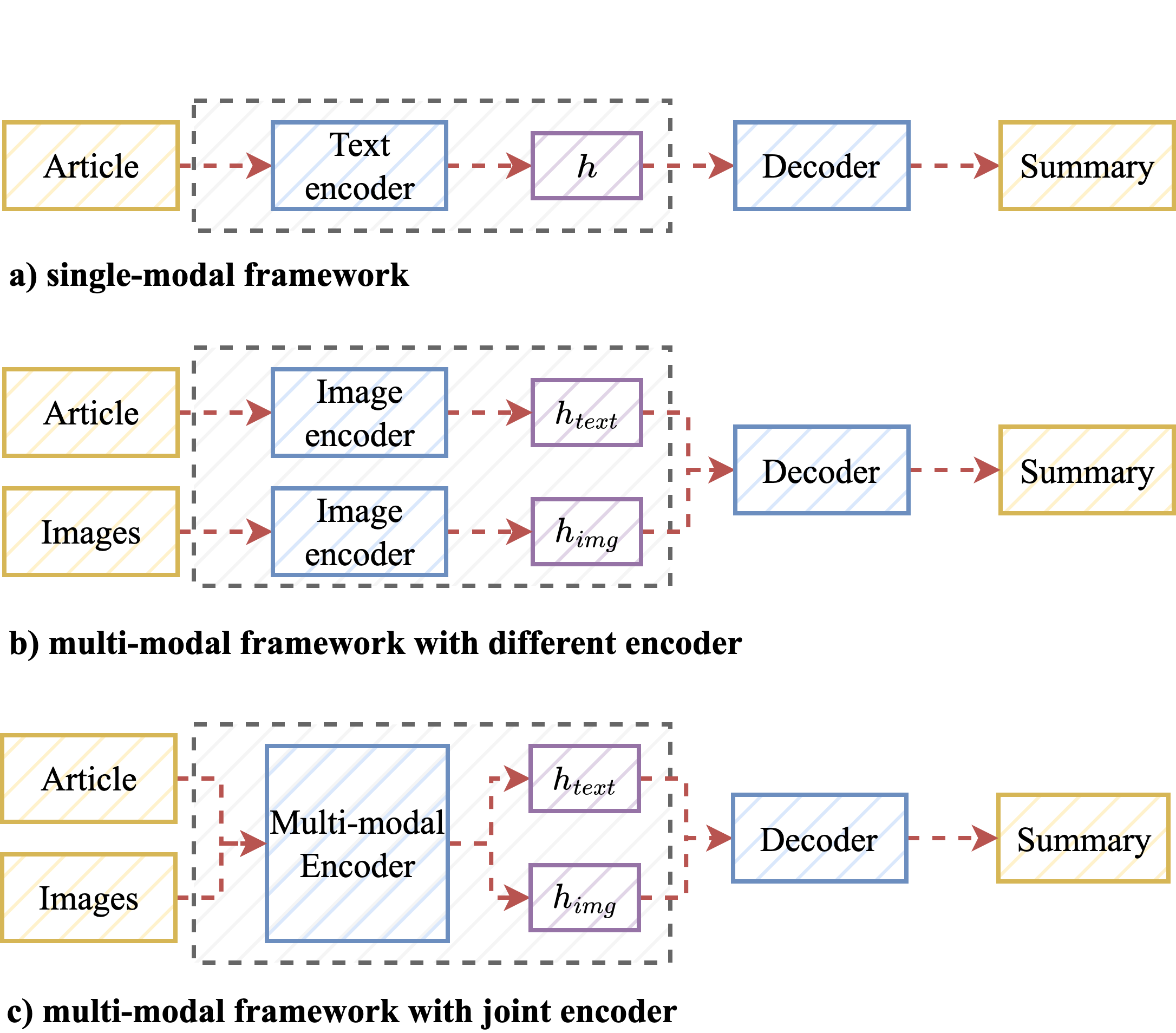}
    \caption{Multi-modal summarization models with different encoder structures.}
    \label{fig:framework}
\end{figure}

Recent single-modal summarization models always employ an encoder-decoder framework with transformers structure~\cite{zhang2019pegasus,lewis-etal-2020-bart}. 
Existing multi-modal models always add separate encoders for different modalities into the single-modal encoder-decoder framework~\cite{li-etal-2017-multi,li-etal-2018-filter,chen-zhuge-2018-abstractive,zhu-etal-2018-msmo,khullar-arora-2020-mast,zhu-etal-2020-guidance,im-etal-2021-self}. We show the widely-used structure of them in Figure~\ref{fig:framework}a and \ref{fig:framework}b. The representation of different modalities is obtained separately from single-modal encoders, which leads to the model can not effectively capture the interaction between them. Recently, some works have paid attention to how to enhance image-text interaction~\cite{zhu-etal-2020-guidance,DBLP:journals/corr/abs-2112-12072} by adding interactive modules or auxiliary tasks.

However, previous works ignored the paragraph-level vision-language semantic alignment, where an example is shown in Figure~\ref{fig:example}. The semantics of each paragraph is highly corresponding to the image on the left.
Besides, visual-language joint encoding is not well-applied for multi-modal summarization tasks, which has been proven effective on many multi-modal natural language understanding (NLU) tasks (e.g. Visual Question Answering)~\cite{DBLP:conf/aaai/LiDFGJ20,li-etal-2020-unicoder,li2020oscar,zhang2021vinvl,xu-etal-2021-e2e,zhou-etal-2020-unified}.

To improve these deficiencies, in this paper, we propose the Vision-Language Summarization model ViL-Sum with a universal transformer-based encoder-decoder structure. The core of ViL-Sum is a joint multi-modal encoder with two well-designed tasks, image reordering, and image selection, which aims to guide the model to learn better vision-language representations and capture the alignment of paragraph-level vision-language semantics. 
Specifically, we use a backbone (e.g. ViT~\cite{dosovitskiy-2021-an}) to convert images into visual token embeddings and concatenate them with document token embeddings as the input of the joint multi-modal encoder. The ViL-Sum structure with the joint multi-modal encoder is shown in Figure~\ref{fig:framework}c. 
To model paragraph-level vision-language semantic alignment, we propose a simple but effective image reordering task. It forces the model to reorder shuffled input images, which guides the model to learn the corresponding relation between paragraphs and images. To further enhance vision-language representation, we also train ViL-Sum with an image selection task, which selects several summary-related images as part of the multi-modal summary. We follow~\cite{zhu-etal-2020-guidance} used image caption construct pseudo labels.
Finally, we train ViL-Sum with text summary generation, image selection, and image reordering tasks in a multi-task manner.

Experiments show that our ViL-Sum with multi-task training can outperform baselines by a wide margin. And further analysis demonstrates that the improvement is exactly from the joint modelling and multi-task training.
However, the caption of the image is not always available. So the image selection task is not generalization for all datasets.
It is deserved to mention that if we remove the image selection task, our proposed multi-modal encoder and the image reordering task still help the model beat all comparison models.

Our contributions can be summarized as follows\footnote{Code will be released after de-anonymous.}:
\begin{itemize}
    \item We proposed a novel vision-language Summarization (ViL-Sum) model, which can jointly encode images and text to capture their interrelation.
    \item We propose two auxiliary tasks and employ multi-task learning to guide the model to learn the paragraph-level vision and language semantic alignment. 
    \item Our model outperforms all current state-of-the-art methods on automatic and manual evaluation metrics. And in further analysis, we find that the improvement is exactly from the paragraph-level semantic alignment modelling and multi-task training.
\end{itemize}

\section{Methodology}

\begin{figure*}[]
    \centering
    \includegraphics[width=0.8\textwidth]{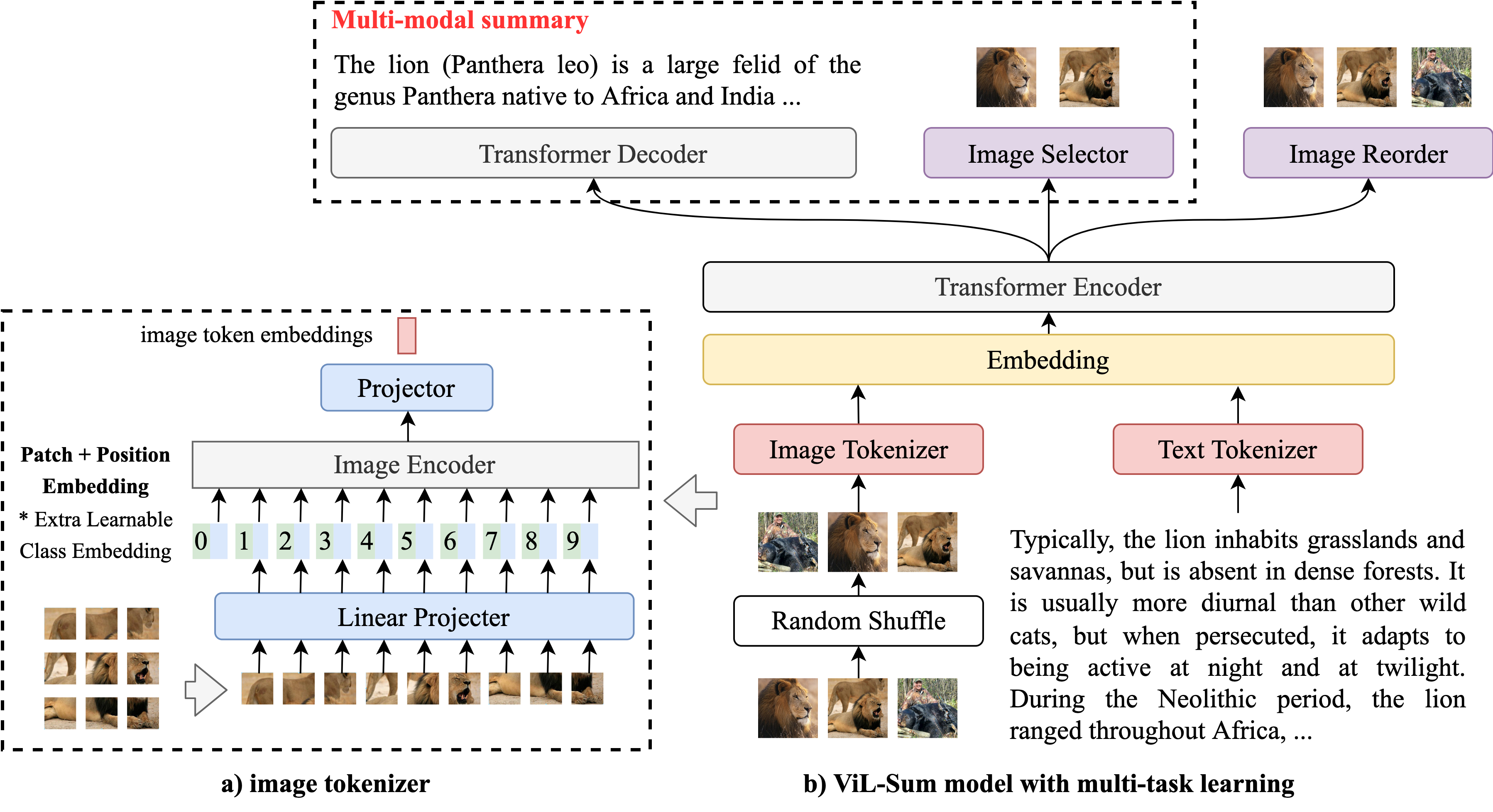}
    \caption{The overall framework of our proposed ViL-Sum model. Figure a) is the detail of the ViT-based image tokenizer. Figure b) is the encoder-decoder framework with multi-task learning.}
    \label{fig:main}
\end{figure*}

We show the main architecture of our ViL-Sum model in Figure~\ref{fig:main}.
Firstly, we employ a backbone network as the image tokenizer to convert images into visual token embeddings in Figure~\ref{fig:main}a. Then, text embeddings and visual token embeddings are concatenated as the input of the main encoder-decoder framework in Figure~\ref{fig:main}b. 
Finally, we train the ViL-Sum model in a multi-task manner. In the following sections, we will first introduce vision-language joint representation. Then, we will describe the details of multi-task learning.

\subsection{Vision-Language Joint Representation}
First of all, we formalized the input and output of our ViL-Sum as $(D, I)$ and $(S, I_S)$, where $D=\{t_1, t_2, \dots, t_T\}$ refers to the sequence of tokens from the input document, $I=\{img_0, img_1, \dots, img_M\}$ refers to the sequence of input images from the input document, $S=\{t_1, t_2, \dots \}$ refers to the sequence of tokens from gold text summary, and $I_S = \{img_1, img_2, \dots, img_K\}$ refers to $K$ selected images for the multi-modal summary. 

\subsubsection{Document Embeddings}
Each document is firstly converted into the sequence of tokens $\{t_1, t_2, \dots, t_T\}$ and then two special tokens ``$\langle s\rangle$" and ``$\langle \backslash s\rangle$" are added to represent the start and end of the document. After that, we map each token into vector representation $E_D = \{e_{start}, e_1, \dots, e_T, e_{end}\}$ with text embedding layer.

\subsubsection{Image Embeddings}
Different from previous methods, which extract many image features via existing object detection models. We employ ViT~\cite{dosovitskiy-2021-an} as the backbone, which split each image into several patches and then encode them. The details of the image tokenizer are shown in Figure~\ref{fig:main}b.

Firstly, we reshape image $img \in \mathbb{R}^{H \times W \times C}$ into a sequence of flattened 2D patches $\{img^p \in \mathbb{R}^{N\times (P^2\cdot C)}\}_{p=1}^N$, where $(H, W)$ is the resolution of the original image, $C$ is the number of channels, $(P, P)$ is the resolution of each image patch, and $N = HW/P^2$ is the resulting number of patches. Then, we can obtain a sequence of image patches $\{img^p\}_{p=1}^N$ as the input of the image tokenizer.

Secondly, the patches are linearly projected to patch embeddings $e^p=E\times img_i^p$, where $E \in \mathbb{R}^{(P^2\cdot D)\times C}$. We also add a special token ``[class]'' with learnable embedding $e^0$. Then attaching position embeddings and patch embeddings as input $Z_0$ for the image encoder to retain positional information of images:
\begin{equation}
    Z_0 = [e_i^0; e_i^1; \dots; e_i^N] + E_{pos}
\end{equation}
Where $Z_0, E_{pos} \in \mathbb{R}^{(N+1)\times D}$, $E_{pos}$ is position embeddings.

Finally, we employ the pre-trained ViT as the backbone to encode these patches of each image. This backbone also can be replaced by any other encoders (e.g. linear projection layer). 
\begin{equation}
    Z_{\ell+1} = \mathtt{Transformer}(Z_{\ell}), \ell=1,2,...,L
\end{equation}
The global max-pooling of output vectors is obtained as the visual token embedding of image $img_i$:
\begin{equation}
    v_i = \mathtt{Maxpooling}(Z_{L})
\end{equation} where $v_i\in {R}^{D}$.
Through the image tokenizer, we can convert the sequence of input images into a sequence of visual token embeddings $E_v=\{v_i\}_{i=1}^{M}$.

\subsubsection{Multi-modal Encoder}
The input of the multi-modal encoder is the concatenation of visual token embeddings $E_v$ and token embeddings $E_D$. We can formalize the input as $H_0 = \{E_v; E_D\}$ and then encode visual and text embeddings with 12 transformer blocks. Finally, we can obtain vision-language representation $H_L$ from the last layer output of this encoder.
\begin{equation}
    H_L = \{h_{v_1},\dots, h_{v_M}, h_{start}, h_{1}, \dots, h_{end}\}
\end{equation}
The vision and language semantics interact with the self-attention mechanism of the transformer structure during the encoding process. 

\subsection{Visual-enhanced Summary Generation}
The vision-language representations $H_L$ from the previous multi-modal encoder contain multi-modal features of input text and images. 
After encoding, we feed the representations $H_L$ into the decoder to generate a text summary. 
The target of the summary generation task is to minimize the negative log-likelihood of the reference $y$ tokens as given input document $D$ and images $I$ via updating model parameters $\theta$. The loss function of the summary generation task is as follows:
\begin{equation}
    \mathcal L_{\theta}^{GEN} = - \sum_{j=1}^{|y|}\log P_{\theta}(y_j|y_{<j}, D, I)
\end{equation}
Different from single-modal summarization tasks, this optimization target also depends on the features from input images $I$, which enhance the final summary generation.

\subsection{Images Reordering}
To align the paragraphs and images from the input, in this section, we introduce a simple yet effective task, image reordering, to guide the model to learn semantic alignment.
Specifically, we shuffle the order of input images and then force the ViL-Sum model to predict the original order of input images with a classification head:
\begin{equation}
y_i=P(pos_i)=\mathtt{softmax}(W\cdot h_{v_i} + b)
\end{equation} where all input images share one classification head.
To train the classification layer, the model computes loss and minimizes the objective function:
\begin{equation}
    \mathcal L_{\theta}^{IR} = \frac{1}{M}\sum_{i=1}^M \sum_{c=1}^C -\hat y_{ic} \log y_{ic}
\end{equation}
where $C$ is the number of categories, depending on the number of input images. We set $C=10$. if the number of input images is greater than 10, we only keep the first 10 images as input images. 

\subsection{Images Selection}
We also train our ViL-Sum with multi-modal output reference following~\cite{zhu-etal-2020-guidance}.
To build pseudo image selection labels of training data, we employ similarity between image caption and gold summary to select top-$K$ images as labels $\hat y$ ($K$ is empirically set as 3). The similarity is the average of ROUGE-1, ROUGE-2, and ROUGE-L scores.
The probability to select each image is as follows:
\begin{equation}
   y_i = P(img_i) = \sigma(W\cdot h_{v_i} + b) 
\end{equation}
The loss function of the image selection task is as follows:
\begin{equation}
    \mathcal L_{\theta}^{IS} = \frac{1}{M} \sum_{i=1}^M -[\hat y_i \log y_i +(1-\hat y_i) \log(1- y_i)]
\end{equation} 

\subsection{Enhanced by Multi-task Learning} \label{sec:sub:mt}
We train our ViL-Sum with a text summary generation task and two well-designed auxiliary tasks in a multi-task manner, which are used to enhance vision-language representation and paragraph-level semantic alignment. In previous sections, we have introduced the details of them.
Finally, ViL-Sum is trained with three tasks: summary generation, image selection, and image reordering, jointly by simultaneously minimizing three loss functions as follows:
\begin{equation}
    \mathcal L_{\theta}^{TOTAL} = L_{\theta}^{GEN} +  L_{\theta}^{IS} + L_{\theta}^{IR} 
\end{equation}
It is deserved to mention that the caption of the image is not always available. So the image selection task is not generalization for all datasets.
If we remove the image selection task, we can select images by measuring the similarity between generated summary and vector representations of images. Our proposed multi-modal encoder and the image reordering task still help the model achieve excellent performance.

\section{Experimental Setup}
\subsection{Dataset}
\begin{table}[]
    \centering
    \caption{Statistical information of MSMO dataset. D refers to the input document. S refers to the summary. }
    \begin{tabular}{l|ccc}
    \toprule \midrule
                & train & valid & test \\ \midrule
    \#Documents  & 293,965 & 10,355 & 10,261 \\
    \#AvgTokens(D) & 721 & 766 & 731\\
    \#AvgTokens(S) & 70 & 70 & 72 \\
    \#Images    & 1,928,356 & 68,520 & 71,509 \\
    \#AvgImgs  & 6.56 & 6.62 & 6.97 \\ \midrule
    \bottomrule
    \end{tabular}
    
    \label{tab:dataset}
\end{table}
We employ the MSMO dataset~\cite{zhu-etal-2018-msmo} to evaluate the effectiveness of our proposed ViL-Sum. MSMO dataset is a large-scale dataset for the Multi-modal Summarization with Multi-modal Output tasks. Each example in the dataset is a triplet (document, images, summary), which contains more than one image in each example. 
This dataset contains online news articles (723 tokens on average) paired with multiple image-caption pairs (6.58 images on average) and multi-sentence summaries (70 tokens on average). For test data, based on text reference, at most three images are annotated to produce a multi-modal reference by humans. The detailed statistical information of the MSMO  dataset is shown in Table~\ref{tab:dataset}.

\subsection{Baseline Models}
We report the existing multi-modal summarization methods (ATG, ATL, HAN, GR)~\cite{zhu-etal-2018-msmo} and MOF$^{RR}_{dec}$~\cite{zhu-etal-2020-guidance} using multiple metrics.
We also report the result of PGC~\cite{see-etal-2017-get}, which is a single-modal summarization model.

To prove the effectiveness of our proposed joint representation and multi-task learning, we mainly compare with BART-base~\cite{lewis-etal-2020-bart} model and a reproduced two-stream model BART-cross which has the same structure with MOF$^{RR}_{dec}$ and replace GRU and VGG19~\cite{liu-2015-vgg} with BART and ViT~\cite{dosovitskiy-2021-an} respectively. To be fair, we mainly compare our model with BART-base and BART-cross due to previous methods did not employ pre-trained models. The details of these models are as follows:

\begin{itemize}
    \item \textbf{PGC} is the BiGRU-based pointer-generator network that allows both copying words from the input text and generating words from a fixed vocabulary.
    \item \textbf{ATG} is based on the PGC model. It fuses static visual features from VGG19 with text features after the BiGRU encoder. Besides, ATG selects final images by the visual-text attention weight.
    \item \textbf{ATL} replaces the image global features of ATG with local features (multiple pooling features), which select images by measuring the sum of visual attention distribution over the local patch features of each image.
    \item \textbf{HAN} is based on the ATL model and adds a hierarchical attention mechanism. This attention mechanism first attends to the image patches to get the intermediate vectors to represent images and then attends to these vectors to get the visual context vector.
    \item \textbf{GR} is an extractive method that employs LexRank~\cite{lexrank} to rank captions of images and select images based on the rank score. The text summary of it is generated by the PGC model.
    \item \textbf{MOF$^{RR}_{dec}$} is based on ATG model. This model first constructs pseudo-labels of image selection for the final summary. Specifically, it employs the ROUGE score to measure the relevance of image caption and summary text. 
    \item \textbf{BART-base} is a pre-trained seq2seq generation model, which achieved promising results in many generations of NLP tasks, especially on text summarization. We employ this model to confirm visual features' contribution to a summary generation.
    \item \textbf{BART-cross} is a BART-based model with the same model structure as previous ATG, ATL, HAN, GR, and MOF$^{RR}_{dec}$. It first encodes images with ViT and then fuses text representation from the BART encoder output. The fusion of image and text representations employs cross-attention like the ATG model. This is the main comparison model.
\end{itemize}
For a fair comparison, we construct this BART-cross model to prove the effectiveness of joint multi-modal encoder and multi-task training in our ViL-Sum. Because our ViL-Sum without multi-task training only changes the encoding mechanism from separate encoders to the joint multi-modal encoder.

\begin{table*}[]
\centering

\begin{tabular}{c|l|ccccc|c} \toprule \midrule
                  & Model                              & ROUGE-1   & ROUGE-2   & ROUGE-L   & MAX$_{sim}$  & IP    & MMAE \\ \midrule
\multirow{6}{*}{1} & PGC (See et al. 2017)                         & 41.11 & 18.31 & 37.74 & -     & -     & -    \\
                  & ATG~\cite{zhu-etal-2018-msmo}                               & 40.63 & 18.12 & 37.53 & 25.82 & 59.28 & 3.35 \\
                  & ATL~\cite{zhu-etal-2018-msmo}                               & 40.86 & 18.27 & 37.75 & 13.26 & 62.44 & 3.26 \\
                  & HAN~\cite{zhu-etal-2018-msmo}                                & 40.82 & 18.30 & 37.70 & 12.22 & 61.83 & 3.25 \\
                  & GR~\cite{zhu-etal-2018-msmo}                                 & 37.13 & 15.03 & 30.21 & \textbf{26.60} & 61.70 & 3.20 \\
                  & MOF$^{RR}_{dec}$~\cite{zhu-etal-2020-guidance}                           & \textbf{41.20} & \textbf{18.33} & \textbf{37.80} & 26.38 & \textbf{65.45} & \textbf{3.37} \\ \midrule
\multirow{5}{*}{2} & BART-base                      & 43.75 & 20.70 & 40.66 & -     & -     & -    \\
                  & BART-cross                         & 43.67 & 20.65 & 40.65 & 30.25 & 65.98 & 3.45 \\
                  & ViL-Sum                       & \textbf{44.29} & \textbf{20.96} & \textbf{41.34} & 32.17 & 66.27 & 3.48 \\
                  & ViL-Sum+SEL                & 44.20 & 20.90 & 41.22 & 34.47 & 68.18 & 3.51 \\
                  & ViL-Sum+REO               & 44.21 & 20.98 & 41.20 & 34.35 & 69.03 & 3.52 \\
                  & ViL-Sum+SEL,REO & 44.16 & 20.88 & 41.21 & \textbf{34.52} & \textbf{71.73} & \textbf{3.55} \\  \midrule
\bottomrule
\end{tabular}
\caption{The main results of all comparison models on different metrics. Models in block 1 are based on the pointer network with Bi-GRU. Models in block 2 are based on the BART-base model. SEL means selection task and REO means reordering task. All reported results of ours are the average of 3 different checkpoints.}
\label{tab:main_res}
\end{table*}

\subsection{Implementation Details}
We train our model for 10 epochs on 8xV100 GPUs using Adam~\cite{kingma-2014-adam} with $\beta_1=0.9$, $\beta_2=0.99$, a batch size of 64. We also use a linear learning rate warm-up with 1,000 steps. The weight-decay is set as $10^{-4}$. 
The model is initialized with ViT-B/16 and BART-base parameters. The max length of input images and tokens are 10 and 512 respectively. 
For the image tokenizer, we employ the same setting with ViT-b/16 in~\cite{dosovitskiy-2021-an}.
During testing, we generate the summary with a beam size of 3, and the minimum and maximum decoding lengths are set as 15 and 150 separately.

\subsection{Evaluation Metrics}
We evaluate the pictorial summary with the MMAE metric~\cite{zhu-etal-2018-msmo}. \footnote{Comment: \newcite{zhu-etal-2020-guidance} also proposed a MMAE+ to better evaluate MSMO task. However, the author did not release their MR model, which is the core component of their MMAE+. We find that the performance of MMAE and MMAE+ is very closer and consistent.}

MMAE consists of three sub-metrics: ROUGE score (ROUGE-L), Image Precision (IP), and Image Text Relevance (MAX$_{sim}$). ROUGE~\cite{lin-2004-rouge} score can measure the salience of text in generated summary, which is widely used for measuring summarization systems. The image precision can measure the salience of selected images and is computed as Equ. (\ref{eq:ip}).
\begin{equation}
    \mathbf{IP} = \frac{|ref_{img} \cap rec_{img}|}{|rec_{img}|}
    \label{eq:ip}
\end{equation}where $ref_{img}$ and $rec_{img}$ denote reference images and
recommended images by MSMO systems respectively. 
MAX$_{sim}$ can measure the relevance between selected images and generated text summary, which trains an image-text retrieval~\cite{faghri-2018-vsepp} model with max-margin loss to evaluate Image-Text relevance.
Finally, Zhu \textit{et. al.}~\cite{zhu-etal-2018-msmo} choose the linear regression results of 3 metrics as MMAE with human judgments and the weight for ROUGE-L, MAX$_{sim}$, and IP is 1.641, 0.854, 0.806 respectively, the intercept is 1.978.

We report the results of ROUGE-1/2/L, MAX$_{sim}$, IP, and MMAE of each model to comprehensively measure their performance. The results of our model are all the averages of three different checkpoints.

\section{Results}

\subsection{Overall Performance}
The main results of all models are shown in Tab. \ref{tab:main_res}. Models in block 1 are based on the pointer network with Bi-GRU. Models in block 2 are based on the BART-base model. SEL means selection task and REO means reordering task. All reported results of ours are the average of 3 different checkpoints.

We can see that compared with the baselines, our ViL-Sum gains significant improvement on all metrics, and ViL-Sum+selection, reordering achieves the best comprehensive performance. Compared with BART-cross, we can see that the joint representation and multi-task training both bring satisfactory improvement, which proved the effectiveness of our proposed methods. Interestingly, the introduction of image features hurts the performance of all single-modal summarization models, especially Bi-GRU-based models.

\subsection{Performance of Joint Representation}
Firstly, we can see that the performance of ATG, ATL, HAN, and GR all hurt ROUGE scores by simply introducing images as independent visual features. Through the multi-modal objective optimization, MOF$^{RR}_{dec}$ has a significant improvement on IP and does not decrease the quality of generated text summary.
This situation proves that modelling vision and language information independently did not bring in the revenue for text summary generation. 
The results of BART-cross, which also introduces images as independent features, also have lower ROUGE scores than BART-base. This situation proved again the previous conclusion.

Different from previous performance on ROUGE score, our ViL-Sum with joint vision-language representation obtains better ROUGE scores, and the Image Precision (IP) and MAX$_sim$ both have a significant improvement. 
This demonstrated that using the joint multi-modal encoder to obtain vision-language representation is better than using separate encoders with cross-attention to fuse multi-modal features.

\subsection{Performance of Multi-task Learning}
The result of ViL-Sum without multi-task learning has achieved new state-of-the-art performance and is better than BART-cross. In this section, we will analyze the influence of our proposed multi-task learning. 
From the results, we can see that the introduction of image selection and reordering bring a slight decrease in ROUGE scores. Meanwhile, the IP and MAX$_{sim}$ scores increase significantly, which makes the overall score MMAE better than ViL-Sum without multi-task training. 

We report the ablation study results of two auxiliary tasks in the second block of Table~\ref{tab:main_res}. From the results, we can see that image selection and reordering both can bring improvement in IP and MAX$_{sim} $ scores. The combination of two tasks can push the overall score MMAE higher.
The comparison of these models demonstrated that the introduction of multi-task learning exactly improved the vision-language representation and semantic alignment, which is reflected in the improvement of the multi-modal metrics: IP, MAX$_{sim}$ and MMAE.

\section{Discussion}
\subsection{Human Evaluation}
\begin{table}
    
    \centering

    \begin{tabular}{l|c}
    \toprule \midrule
    Systems & Human Score \\
    \midrule
    BART-base   & 3.29\\
    BART-cross  & 3.46\\
    ViL-Sum (best) & \textbf{3.78}\\
    \midrule
    Reference    & \textbf{4.02} \\ \midrule
    \bottomrule
    \end{tabular}
    \caption{Results evaluated by human annotators. Each summary is scored by three persons, and we take the average value.}
    \label{tab:human}
\end{table}

We randomly sample 100 examples from the test set to conduct the human evaluation. The multi-modal summary of golden reference, BART-base, BART-cross, and our ViL-Sum (best) are evaluated by three human annotators. Each annotator will score each example with a rating scale from 1 (worst) to 5 (best). 
Table~\ref{tab:human} shows the average scores from three annotators (t-test, $p<0.05$). We can see that annotators tend to give the multi-modal summary from BART-cross and our ViL-Sum higher scores. In addition, our ViL-Sum outperforms two strong baselines by a wide margin and is close to the references.

\subsection{Impact of Different Numbers of Images}
\begin{table}
    \centering
    \small
    
    \begin{tabular}{l|ccc|c}
    \toprule \midrule
    $K$ & ROUGE-L & MAX$_{sim}$ & IP & MMAE \\
    \midrule
    1     & 40.97 & 34.63 & 70.94 & 3.54 \\
    2     & 41.12 & 34.33 & 70.40 & 3.53 \\
    3     & \textbf{41.21} & \textbf{34.52} & \textbf{71.73} & \textbf{3.55} \\
    4     & 41.08 & 34.49 & 70.61 &  3.53 \\ \midrule
    \bottomrule
    \end{tabular}
    \caption{Results of ViL-Sum under different numbers $K$ of images, which are selected into the final summary.}
    \label{tab:k}
\end{table}
Tab.~\ref{tab:k} depicts the experimental results of our model performance varying with different $K$ (the image number at the final summary). Since the golden reference in the test set contains three images, the consistency between training and test makes the model perform best when $K$ is 3. Overall, our model is not very sensitive with $K$. With different $K$, our ViL-Sum all achieve excellent performance, which proves our method can identify the real importance images from multi-modal inputs. Besides, we guess the image selection of the MSMO dataset is simple due to the data from the news.

\subsection{Impact of Different Image Tokenizer}
\begin{table}[]
    \small
    \centering
  
    \begin{tabular}{l|ccc|c}
    \toprule \midrule
    & ROUGE-L & MAX$_{sim}$ & IP & MMAE \\
    \midrule
    ViT        & \textbf{41.21} & \textbf{34.52} & \textbf{71.73} & \textbf{3.55} \\
    \midrule
    Linear     & 40.18 & 33.89 & 70.44 & 3.51 \\
    Vision     & 41.10 & 34.28 & 71.04 & 3.54 \\ \midrule
    \bottomrule
    \end{tabular}
      \caption{Results of ViL-Sum with different image tokenizers. Linear means image tokenizer which replaces transformer blocks with linear layers. Vision is an image tokenizer from Vision Transformer.}
    \label{tab:tok}
\end{table}

To further evaluate the effectiveness of joint modelling and multi-task learning, we replace the backbone of the image tokenizer to observe the performance of ViL-Sum. We replace the ViT backbone with Linear Layer and an image tokenizer from Vision Transformer~\cite{wu2020visual}. Both of them have much smaller parameters than the ViT backbone. Specifically, linear is the simple version of ViT which replaces the transformer image encoder with a simple linear layer to map the images into visual token embeddings. Vision is an image tokenizer from Vision Transformer~\cite{wu2020visual}, which can convert one image into several visual token embeddings.
Table~\ref{tab:tok} reports the results of them. We can see that the ViT exactly provides better visual features than the other two backbones. However, the performance does not drop sharply with the replacement of the image tokenizer. This proves that Our proposed two strategies are robust and the ViL-Sum is flexible with different image tokenizers.

\begin{figure*}[ht]
    \centering
    \includegraphics[width=\textwidth]{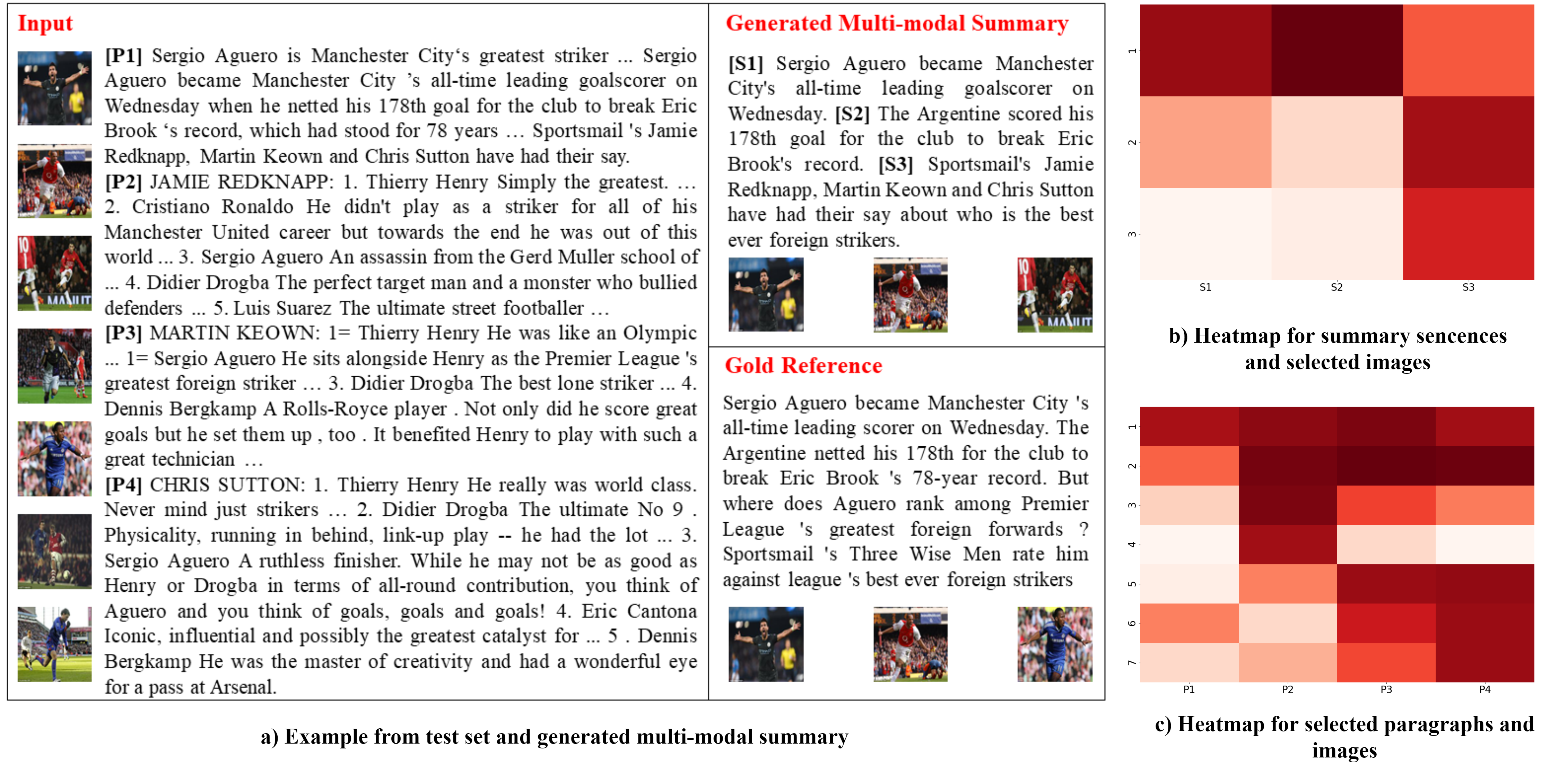}
    \caption{Example from the test set with the generated multi-modal summary. Figure a) is the full example. Figure b) is the heatmap that shows the relevance of the summary and selected images. Figure c) is the heatmap that shows the relevance of selected paragraphs and images. Each color block means cosine similarity between the image and text object. The darker color refers to higher similarity.}
    \label{fig:case}
\end{figure*}

\begin{figure}
    \centering
    \includegraphics[width=0.45\textwidth]{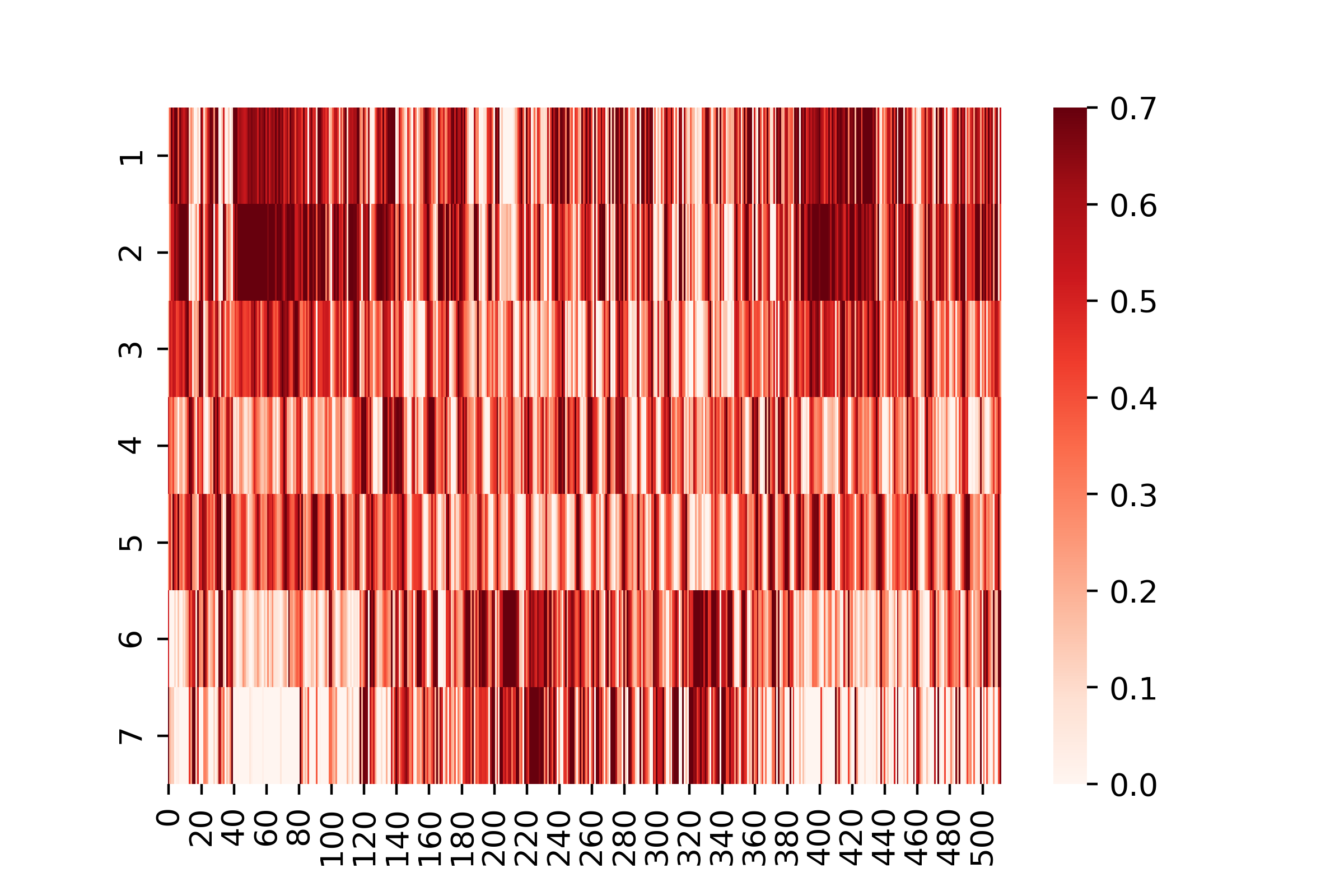}
    \caption{The heatmap shows the relevance of all input tokens and images. The darker color refers to higher similarity.}
    \label{fig:case2}
\end{figure}

\subsection{Case Study and Relevance Visualization}

We select one typical example from the test set and visualize the relevance of 1) summary sentences and selected images; 2) selected paragraphs and images; 3) all tokens and images in Figure~\ref{fig:case} and \ref{fig:case2}. 
Each colour block means a cosine similarity between the image and text object. The darker colour refers to a higher similarity in the heatmap. 
With our proposed methods, the generated summary contains high-quality summary with three related images as shown in Figure~\ref{fig:case}a.
From three different relevant visualizations, we can see that our ViL-Sum can effectively align the semantic representation of summary sentences and selected images as shown in Figure~\ref{fig:case}b. 
The input images can be aligned with paragraphs by training with image reordering as shown in Figure~\ref{fig:case}c.
We also report the heatmap of all input tokens and images in Figure ~\ref{fig:case2}, which is consistent with Figure~\ref{fig:case}b and \ref{fig:case}c. This case proves that the multi-task training really helps ViL-Sum learn reasonable relations between images and input paragraphs.

\section{Related Work}

\subsection{Single-modal Summarization}
Recently, text summarization models have achieved remarkable performance with the development of pre-trained language models. Liu and Lapata~\cite{liu-lapata-2019-text} first apply the pre-trained language model BERT to summarization tasks. They added several transformers as the decoder to the BERT encoder and then train them with different learning rates. Their work outperformed all traditionally trained neural models. Pegasus~\cite{zhang2019pegasus} and BART~\cite{lewis-etal-2020-bart} are two fully pre-trained models for summarization generation with well-designed self-supervised tasks. Their appearance provided powerful base models for summarization and totally changed the research paradigm in the summarization task. After that, more and more summarization works begin to focus on pre-trained language models, including supervised and unsupervised methods~\cite{zhong-etal-2020-extractive,liang-etal-2021-improving,9664266,liang-etal-2022-efficient,DBLP:journals/csur/DongLGCLSY23}.

 \subsection{Vision-Language Representation}
Large-scale Transformers-based~\cite{vaswani-etal-2017-attention} vision and language representation models~\cite{radford2019language,devlin2019bert,lewis-etal-2020-bart} have achieved state-of-the-art results on many Natural Language Processing (NLP) tasks. They first pre-trained on a large-scale corpus with self-supervised tasks and then fine-tuned on specific downstream tasks. 
Most existing vision and language pretraining (VLP) models~\cite{tan2019lxmert,li2021semvlp} adopt two different encoders to model vision and language separately, which extracts visual features by an object detection model and then combines the derived object-centric representation of the image and text.  
Recently, large-scale vision and language representation learning has tried to jointly encode different modalities with the same encoder and achieved promising improvements~\cite{DBLP:conf/aaai/LiDFGJ20,li-etal-2020-unicoder,li2020oscar,zhang2021vinvl,xu-etal-2021-e2e,zhou-etal-2020-unified}.
Their success proves the joint modelling of different modalities is practicable.

\subsection{Multi-modal Summarization}
Different from single-modal text summarization, multi-modal summarization is a task to generate a condensed summary to cover the primary information from multimedia data.
One of the most significant characteristics of this task is it is not only based on text information, but can also employ rich visual information from images, audio, and videos.
Multi-modal summarization tasks can be divided into two types with different outputs: single-modal output~\cite{6527322,chen-zhuge-2018-abstractive,li-etal-2018-filter} and multi-modal output~\cite{bian-etal-2015-microblog,zhu-etal-2018-msmo,zhu-etal-2020-guidance}.
Compared with single-modal output, a multi-modal output summary can increase users’ satisfaction~\cite{zhu-etal-2018-msmo} and first proposed a large-scale Multi-modal Summarization with Multi-modal Output (MSMO) dataset.
To tackle the gap between training and testing in the MSMO task, Zhu \textit{et. al.}~\cite{zhu-etal-2020-guidance} proposed two methods to obtain pseudo image labels and train the model with multi-modal optimization objectives.

However, previous works all obtain vision-language representation via separate encoders for different modalities, which has been proved weaker than joint representation in vision-language representation learning research~\cite{zhou-etal-2020-unified,xu-etal-2021-e2e}. Besides, they ignored the special paragraph-level semantic alignment between different modalities. In this paper, we proposed a novel vision-language summarization ViL-Sum model with a multi-task learning framework to tackle these issues.

\section{Conclusion}
In this paper, we proposed a novel Vision-Language Summarization (ViL-Sum) model, which can enhance the vision-language representation and the paragraph-level semantics alignment through multi-task training and joint modelling. Our ViL-Sum achieves new state-of-the-art results on automatic and manual evaluation metrics. Further analysis demonstrates the improvement is from the joint multi-modal encoder and multi-task training.
Our proposed image reordering task is straightforward yet effective. We believe it can be extended to more scenarios (e.g. vision-language pre-training models) and modalities (e.g. audio and video).

\bibliography{anthology,custom}

\begin{thebibliography}{37}
\expandafter\ifx\csname natexlab\endcsname\relax\def\natexlab#1{#1}\fi

\bibitem[{Bian et~al.(2015)Bian, Yang, Zhang, and
  Chua}]{bian-etal-2015-microblog}
Jingwen Bian, Yang Yang, Hanwang Zhang, and Tat-Seng Chua. 2015.
\newblock \href {https://doi.org/10.1109/TMM.2014.2384912} {Multimedia
  summarization for social events in microblog stream}.
\newblock \emph{IEEE Transactions on Multimedia}, 17(2):216--228.

\bibitem[{Chen and Zhuge(2018)}]{chen-zhuge-2018-abstractive}
Jingqiang Chen and Hai Zhuge. 2018.
\newblock \href {https://doi.org/10.18653/v1/D18-1438} {Abstractive text-image
  summarization using multi-modal attentional hierarchical {RNN}}.
\newblock In \emph{EMNLP}, pages 4046--4056, Brussels, Belgium. Association for
  Computational Linguistics.

\bibitem[{Devlin et~al.(2019)Devlin, Chang, Lee, and
  Toutanova}]{devlin2019bert}
Jacob Devlin, Ming-Wei Chang, Kenton Lee, and Kristina Toutanova. 2019.
\newblock \href {http://arxiv.org/abs/1810.04805} {Bert: Pre-training of deep
  bidirectional transformers for language understanding}.

\bibitem[{Dong et~al.(2023)Dong, Li, Gong, Chen, Li, Shen, and
  Yang}]{DBLP:journals/csur/DongLGCLSY23}
Chenhe Dong, Yinghui Li, Haifan Gong, Miaoxin Chen, Junxin Li, Ying Shen, and
  Min Yang. 2023.
\newblock \href {https://doi.org/10.1145/3554727} {A survey of natural language
  generation}.
\newblock \emph{{ACM} Comput. Surv.}, 55(8):173:1--173:38.

\bibitem[{Dosovitskiy et~al.(2021)Dosovitskiy, Beyer, Kolesnikov, Weissenborn,
  Zhai, Unterthiner, Dehghani, Minderer, Heigold, Gelly, Uszkoreit, and
  Houlsby}]{dosovitskiy-2021-an}
Alexey Dosovitskiy, Lucas Beyer, Alexander Kolesnikov, Dirk Weissenborn,
  Xiaohua Zhai, Thomas Unterthiner, Mostafa Dehghani, Matthias Minderer, Georg
  Heigold, Sylvain Gelly, Jakob Uszkoreit, and Neil Houlsby. 2021.
\newblock \href {https://openreview.net/forum?id=YicbFdNTTy} {An image is worth
  16x16 words: Transformers for image recognition at scale}.
\newblock In \emph{International Conference on Learning Representations}.

\bibitem[{Erkan and Radev(2004)}]{lexrank}
G\"{u}nes Erkan and Dragomir~R. Radev. 2004.
\newblock Lexrank: Graph-based lexical centrality as salience in text
  summarization.
\newblock \emph{J. Artif. Int. Res.}, 22(1):457–479.

\bibitem[{Evangelopoulos et~al.(2013)Evangelopoulos, Zlatintsi, Potamianos,
  Maragos, Rapantzikos, Skoumas, and Avrithis}]{6527322}
Georgios Evangelopoulos, Athanasia Zlatintsi, Alexandros Potamianos, Petros
  Maragos, Konstantinos Rapantzikos, Georgios Skoumas, and Yannis Avrithis.
  2013.
\newblock \href {https://doi.org/10.1109/TMM.2013.2267205} {Multimodal saliency
  and fusion for movie summarization based on aural, visual, and textual
  attention}.
\newblock \emph{IEEE Transactions on Multimedia}, 15(7):1553--1568.

\bibitem[{Faghri et~al.(2018)Faghri, Fleet, Kiros, and
  Fidler}]{faghri-2018-vsepp}
Fartash Faghri, David~J. Fleet, J.~Kiros, and S.~Fidler. 2018.
\newblock Vse++: Improving visual-semantic embeddings with hard negatives.
\newblock In \emph{BMVC}.

\bibitem[{Im et~al.(2021)Im, Kim, Lee, Cho, and Chung}]{im-etal-2021-self}
Jinbae Im, Moonki Kim, Hoyeop Lee, Hyunsouk Cho, and Sehee Chung. 2021.
\newblock \href {https://doi.org/10.18653/v1/2021.acl-long.33} {Self-supervised
  multimodal opinion summarization}.
\newblock In \emph{Proceedings of the 59th Annual Meeting of the Association
  for Computational Linguistics and the 11th International Joint Conference on
  Natural Language Processing (Volume 1: Long Papers)}, pages 388--403, Online.
  Association for Computational Linguistics.

\bibitem[{Khullar and Arora(2020)}]{khullar-arora-2020-mast}
Aman Khullar and Udit Arora. 2020.
\newblock \href {https://doi.org/10.18653/v1/2020.nlpbt-1.7} {{MAST}:
  Multimodal abstractive summarization with trimodal hierarchical attention}.
\newblock In \emph{Proceedings of the First International Workshop on Natural
  Language Processing Beyond Text}, pages 60--69, Online. Association for
  Computational Linguistics.

\bibitem[{Kingma and Ba(2015)}]{kingma-2014-adam}
Diederik~P Kingma and Jimmy Ba. 2015.
\newblock Adam: A method for stochastic optimization.
\newblock In \emph{ICLR 2015}.

\bibitem[{Lewis et~al.(2020)Lewis, Liu, Goyal, Ghazvininejad, Mohamed, Levy,
  Stoyanov, and Zettlemoyer}]{lewis-etal-2020-bart}
Mike Lewis, Yinhan Liu, Naman Goyal, Marjan Ghazvininejad, Abdelrahman Mohamed,
  Omer Levy, Veselin Stoyanov, and Luke Zettlemoyer. 2020.
\newblock \href {https://doi.org/10.18653/v1/2020.acl-main.703} {{BART}:
  Denoising sequence-to-sequence pre-training for natural language generation,
  translation, and comprehension}.
\newblock In \emph{ACL 2020}, pages 7871--7880, Online. Association for
  Computational Linguistics.

\bibitem[{Li et~al.(2021)Li, Yan, Xu, Luo, Wang, Bi, and Huang}]{li2021semvlp}
Chenliang Li, Ming Yan, Haiyang Xu, Fuli Luo, Wei Wang, Bin Bi, and Songfang
  Huang. 2021.
\newblock \href {http://arxiv.org/abs/2103.07829} {Semvlp: Vision-language
  pre-training by aligning semantics at multiple levels}.

\bibitem[{Li et~al.(2020{\natexlab{a}})Li, Duan, Fang, Gong, and
  Jiang}]{DBLP:conf/aaai/LiDFGJ20}
Gen Li, Nan Duan, Yuejian Fang, Ming Gong, and Daxin Jiang. 2020{\natexlab{a}}.
\newblock Unicoder-vl: {A} universal encoder for vision and language by
  cross-modal pre-training.
\newblock In \emph{AAAI 2020}, pages 11336--11344. {AAAI} Press.

\bibitem[{Li et~al.(2020{\natexlab{b}})Li, Duan, Fang, Gong, and
  Jiang}]{li-etal-2020-unicoder}
Gen Li, Nan Duan, Yuejian Fang, Ming Gong, and Daxin Jiang. 2020{\natexlab{b}}.
\newblock \href {https://doi.org/10.1609/aaai.v34i07.6795} {Unicoder-vl: A
  universal encoder for vision and language by cross-modal pre-training}.
\newblock \emph{Proceedings of the AAAI Conference on Artificial Intelligence},
  34(07):11336--11344.

\bibitem[{Li et~al.(2018)Li, Zhu, Liu, Zhang, and Zong}]{li-etal-2018-filter}
Haoran Li, Junnan Zhu, Tianshang Liu, Jiajun Zhang, and Chengqing Zong. 2018.
\newblock \href {https://doi.org/10.24963/ijcai.2018/577} {Multi-modal sentence
  summarization with modality attention and image filtering}.
\newblock In \emph{IJCAI-18}, pages 4152--4158. International Joint Conferences
  on Artificial Intelligence Organization.

\bibitem[{Li et~al.(2017)Li, Zhu, Ma, Zhang, and Zong}]{li-etal-2017-multi}
Haoran Li, Junnan Zhu, Cong Ma, Jiajun Zhang, and Chengqing Zong. 2017.
\newblock \href {https://doi.org/10.18653/v1/D17-1114} {Multi-modal
  summarization for asynchronous collection of text, image, audio and video}.
\newblock In \emph{EMNLP 2017}, pages 1092--1102, Copenhagen, Denmark.
  Association for Computational Linguistics.

\bibitem[{Li et~al.(2020{\natexlab{c}})Li, Yin, Li, Hu, Zhang, Zhang, Wang, Hu,
  Dong, Wei, Choi, and Gao}]{li2020oscar}
Xiujun Li, Xi~Yin, Chunyuan Li, Xiaowei Hu, Pengchuan Zhang, Lei Zhang, Lijuan
  Wang, Houdong Hu, Li~Dong, Furu Wei, Yejin Choi, and Jianfeng Gao.
  2020{\natexlab{c}}.
\newblock Oscar: Object-semantics aligned pre-training for vision-language
  tasks.
\newblock \emph{ECCV 2020}.

\bibitem[{Liang et~al.(2021{\natexlab{a}})Liang, Li, Wu, Li, and Li}]{9664266}
Xinnian Liang, Jing Li, Shuangzhi Wu, Mu~Li, and Zhoujun Li.
  2021{\natexlab{a}}.
\newblock \href {https://doi.org/10.1109/TASLP.2021.3138673} {Improving
  unsupervised extractive summarization by jointly modeling facet and
  redundancy}.
\newblock \emph{IEEE/ACM Transactions on Audio, Speech, and Language
  Processing}, pages 1--1.

\bibitem[{Liang et~al.(2022)Liang, Li, Wu, Zeng, Jiang, Li, and
  Li}]{liang-etal-2022-efficient}
Xinnian Liang, Jing Li, Shuangzhi Wu, Jiali Zeng, Yufan Jiang, Mu~Li, and
  Zhoujun Li. 2022.
\newblock \href {https://aclanthology.org/2022.coling-1.558} {An efficient
  coarse-to-fine facet-aware unsupervised summarization framework based on
  semantic blocks}.
\newblock In \emph{Proceedings of the 29th International Conference on
  Computational Linguistics}, pages 6415--6425, Gyeongju, Republic of Korea.
  International Committee on Computational Linguistics.

\bibitem[{Liang et~al.(2021{\natexlab{b}})Liang, Wu, Li, and
  Li}]{liang-etal-2021-improving}
Xinnian Liang, Shuangzhi Wu, Mu~Li, and Zhoujun Li. 2021{\natexlab{b}}.
\newblock \href {https://doi.org/10.18653/v1/2021.findings-acl.147} {Improving
  unsupervised extractive summarization with facet-aware modeling}.
\newblock In \emph{Findings of the Association for Computational Linguistics:
  ACL-IJCNLP 2021}, pages 1685--1697, Online. Association for Computational
  Linguistics.

\bibitem[{Lin(2004)}]{lin-2004-rouge}
Chin-Yew Lin. 2004.
\newblock \href {https://aclanthology.org/W04-1013} {{ROUGE}: A package for
  automatic evaluation of summaries}.
\newblock In \emph{Text Summarization Branches Out}, pages 74--81, Barcelona,
  Spain. Association for Computational Linguistics.

\bibitem[{Liu and Deng(2015)}]{liu-2015-vgg}
Shuying Liu and Weihong Deng. 2015.
\newblock \href {https://doi.org/10.1109/ACPR.2015.7486599} {Very deep
  convolutional neural network based image classification using small training
  sample size}.
\newblock In \emph{2015 3rd IAPR Asian Conference on Pattern Recognition
  (ACPR)}, pages 730--734.

\bibitem[{Liu and Lapata(2019)}]{liu-lapata-2019-text}
Yang Liu and Mirella Lapata. 2019.
\newblock \href {https://doi.org/10.18653/v1/D19-1387} {Text summarization with
  pretrained encoders}.
\newblock In \emph{EMNLP-IJCNLP 2019}, pages 3730--3740, Hong Kong, China.
  Association for Computational Linguistics.

\bibitem[{Radford et~al.(2019)Radford, Wu, Child, Luan, Amodei, and
  Sutskever}]{radford2019language}
Alec Radford, Jeff Wu, Rewon Child, David Luan, Dario Amodei, and Ilya
  Sutskever. 2019.
\newblock Language models are unsupervised multitask learners.

\bibitem[{See et~al.(2017)See, Liu, and Manning}]{see-etal-2017-get}
Abigail See, Peter~J. Liu, and Christopher~D. Manning. 2017.
\newblock \href {https://doi.org/10.18653/v1/P17-1099} {Get to the point:
  Summarization with pointer-generator networks}.
\newblock In \emph{Proceedings of the 55th Annual Meeting of the Association
  for Computational Linguistics (Volume 1: Long Papers)}, pages 1073--1083,
  Vancouver, Canada. Association for Computational Linguistics.

\bibitem[{Tan and Bansal(2019)}]{tan2019lxmert}
Hao Tan and Mohit Bansal. 2019.
\newblock Lxmert: Learning cross-modality encoder representations from
  transformers.
\newblock In \emph{Proceedings of the 2019 Conference on Empirical Methods in
  Natural Language Processing}.

\bibitem[{Vaswani et~al.(2017)Vaswani, Shazeer, Parmar, Uszkoreit, Jones,
  Gomez, Kaiser, and Polosukhin}]{vaswani-etal-2017-attention}
Ashish Vaswani, Noam Shazeer, Niki Parmar, Jakob Uszkoreit, Llion Jones,
  Aidan~N. Gomez, undefinedukasz Kaiser, and Illia Polosukhin. 2017.
\newblock Attention is all you need.
\newblock In \emph{Proceedings of the 31st International Conference on Neural
  Information Processing Systems}, NIPS'17, page 6000–6010, Red Hook, NY,
  USA. Curran Associates Inc.

\bibitem[{Wu et~al.(2020)Wu, Xu, Dai, Wan, Zhang, Yan, Tomizuka, Gonzalez,
  Keutzer, and Vajda}]{wu2020visual}
Bichen Wu, Chenfeng Xu, Xiaoliang Dai, Alvin Wan, Peizhao Zhang, Zhicheng Yan,
  Masayoshi Tomizuka, Joseph Gonzalez, Kurt Keutzer, and Peter Vajda. 2020.
\newblock \href {http://arxiv.org/abs/2006.03677} {Visual transformers:
  Token-based image representation and processing for computer vision}.

\bibitem[{Xu et~al.(2021)Xu, Yan, Li, Bi, Huang, Xiao, and
  Huang}]{xu-etal-2021-e2e}
Haiyang Xu, Ming Yan, Chenliang Li, Bin Bi, Songfang Huang, Wenming Xiao, and
  Fei Huang. 2021.
\newblock \href {https://doi.org/10.18653/v1/2021.acl-long.42} {{E}2{E}-{VLP}:
  End-to-end vision-language pre-training enhanced by visual learning}.
\newblock In \emph{Proceedings of the 59th Annual Meeting of the Association
  for Computational Linguistics and the 11th International Joint Conference on
  Natural Language Processing (Volume 1: Long Papers)}, pages 503--513, Online.
  Association for Computational Linguistics.

\bibitem[{Zhang et~al.(2019)Zhang, Zhao, Saleh, and Liu}]{zhang2019pegasus}
Jingqing Zhang, Yao Zhao, Mohammad Saleh, and Peter~J. Liu. 2019.
\newblock \href {http://arxiv.org/abs/1912.08777} {Pegasus: Pre-training with
  extracted gap-sentences for abstractive summarization}.

\bibitem[{Zhang et~al.(2021{\natexlab{a}})Zhang, Zhang, Pan, and
  Huang}]{DBLP:journals/corr/abs-2112-12072}
Litian Zhang, Xiaoming Zhang, Junshu Pan, and Feiran Huang. 2021{\natexlab{a}}.
\newblock \href {http://arxiv.org/abs/2112.12072} {Hierarchical cross-modality
  semantic correlation learning model for multimodal summarization}.
\newblock \emph{CoRR}, abs/2112.12072.

\bibitem[{Zhang et~al.(2021{\natexlab{b}})Zhang, Li, Hu, Yang, Zhang, Wang,
  Choi, and Gao}]{zhang2021vinvl}
Pengchuan Zhang, Xiujun Li, Xiaowei Hu, Jianwei Yang, Lei Zhang, Lijuan Wang,
  Yejin Choi, and Jianfeng Gao. 2021{\natexlab{b}}.
\newblock Vinvl: Making visual representations matter in vision-language
  models.
\newblock \emph{CVPR 2021}.

\bibitem[{Zhong et~al.(2020)Zhong, Liu, Chen, Wang, Qiu, and
  Huang}]{zhong-etal-2020-extractive}
Ming Zhong, Pengfei Liu, Yiran Chen, Danqing Wang, Xipeng Qiu, and Xuanjing
  Huang. 2020.
\newblock \href {https://doi.org/10.18653/v1/2020.acl-main.552} {Extractive
  summarization as text matching}.
\newblock In \emph{ACL 2020}, pages 6197--6208, Online. Association for
  Computational Linguistics.

\bibitem[{Zhou et~al.(2020)Zhou, Palangi, Zhang, Hu, Corso, and
  Gao}]{zhou-etal-2020-unified}
Luowei Zhou, Hamid Palangi, Lei Zhang, Houdong Hu, Jason Corso, and Jianfeng
  Gao. 2020.
\newblock \href {https://doi.org/10.1609/aaai.v34i07.7005} {Unified
  vision-language pre-training for image captioning and vqa}.
\newblock \emph{Proceedings of the AAAI Conference on Artificial Intelligence},
  34(07):13041--13049.

\bibitem[{Zhu et~al.(2018)Zhu, Li, Liu, Zhou, Zhang, and
  Zong}]{zhu-etal-2018-msmo}
Junnan Zhu, Haoran Li, Tianshang Liu, Yu~Zhou, Jiajun Zhang, and Chengqing
  Zong. 2018.
\newblock \href {https://doi.org/10.18653/v1/D18-1448} {{MSMO}: Multimodal
  summarization with multimodal output}.
\newblock In \emph{Proceedings of the 2018 Conference on Empirical Methods in
  Natural Language Processing}, pages 4154--4164, Brussels, Belgium.
  Association for Computational Linguistics.

\bibitem[{Zhu et~al.(2020)Zhu, Zhou, Zhang, Li, Zong, and
  Li}]{zhu-etal-2020-guidance}
Junnan Zhu, Yu~Zhou, Jiajun Zhang, Haoran Li, Chengqing Zong, and Changliang
  Li. 2020.
\newblock \href {https://doi.org/10.1609/aaai.v34i05.6525} {Multimodal
  summarization with guidance of multimodal reference}.
\newblock \emph{Proceedings of the AAAI Conference on Artificial Intelligence},
  34(05):9749--9756.

\end{thebibliography}
\bibliographystyle{acl_natbib}




\end{document}